%% file: main.tex
\documentclass[letterpaper]{article}
\usepackage{conf}
\usepackage{times}
\usepackage{helvet}
\usepackage{courier}

\usepackage{amsmath}
\usepackage{amsfonts}
\usepackage{amsthm}
\usepackage{amssymb}
\usepackage{tikz}
\usepackage{dsfont}
\usepackage{subcaption}

\begin{document}
\title{Non parametric estimation of causal populations in a counterfactual scenario}

\author{Celine Beji, Florian Yger, Jamal Atif\\Dauphine-PSL University}

\maketitle
\begin{abstract}
In causality, estimating the effect of a treatment without confounding inference remains a major issue because requires to assess the outcome in both case with and without treatment.
Not being able to observe simultaneously both of them, the estimation of potential outcome remains a challenging task.
We propose an innovative approach where the problem is reformulated as a missing data model. The aim is to estimate the hidden distribution of \emph{causal populations}, defined as a function of treatment and outcome. A Causal Auto-Encoder (CAE), enhanced by a prior dependent on treatment and outcome information, assimilates the latent space to the probability distribution of the target populations.
The features are reconstructed after being reduced to a latent space and constrained by a mask introduced in the intermediate layer of the network, containing treatment and outcome information.
\end{abstract}

\section{Introduction}

This paper adopts the framework of Rubin-Neyman~\cite{Rubin_1974,rubin1975bayesian}, considering causal inference as a missing data problem where the confounding bias is tackled by the estimation of the counterfactual outcome on each value of treatment. For example in healthcare, where the treatment is an appropriate medication and the intended effect is the recovery of the patient, the causal effect is inferred by the potential outcome which is defined as the set of outcomes with and without treatment. The fundamental problem is that only one outcome can be observed, the other is missing.
Once the potential outcome is predicted, an identification of the individuals who respond to the treatment according to their fractures like age, weight, medical history and monitoring indicators, can be made. 
In this context, four \emph{causal populations} are defined: (i) \emph{responders} (R) who only achieve the intended effect if they are treated; (ii) \emph{doomed} (D) who never achieve the intended effect; (iii) \emph{survivor} (S) who always achieve the intended effect; and (vi) \emph{anti-responders} (A) who achieve the intended effect only if they are untreated and for whom the treatment should be unassigned.
The aim of this paper is to estimate the probability distribution of these four distinct causal populations, under compliance, i.e. the assigned treatment is taken. The counterfactual outcome is therefore predicted by assuming that individuals belong to the group with the most significant probability. Although the approach is generalized in multi-treatment, the standard binary treatment assumption is considered. 

Using meta-learner classification~\cite{kunzel2019metalearners}, related work can be classified in three main approaches. First, T-learners approaches refer to algorithms with ``two'' separate models for the treated and untreated individuals. Two different distributions for each treatment group are estimated and the bias between the distributions is reduced by creating subgroups~\cite{foster2011subgroup,lu2018estimating}, by minimizing the distance between the distributions and by using a representation learning~\cite{shalit2017estimating}, a function with two outputs with multi-task Gaussian processes~\cite{alaa2017bayesian}, a deep counterfactual network using a propensity-dropout~\cite{alaa2017deep}. Second, S-learners consist to estimate a ``Single'' response function and share information across estimators of treated and untreated individuals. Many of the models proposed in recent years belong to this category, particularly those with neural network architecture for which the treatment is given as input. In~\cite{hill2011bayesian}, the estimator is modeled with an additive regression tree, in~\cite{johansson2016learning} the problem is reformulated as a domain adaptation problem and a learning representation is used, and in~\cite{yoon2018ganite} a GAN architecture is proposed.
Finally, indirect approaches estimate directly the individual treatment effect without estimating the counterfactual outcome. This is used in~\cite{hitsch2018heterogeneous} which transforms the outcome with an inverse probability weighting based on the propensity score, also in~\cite{kang2007demystifying,knaus2021machine} the authors use a double robust estimation and for the R-learner in~\cite{nie2020quasi} the authors have modified the loss function.

In contrast to related work, estimating the probability distribution of causal populations provides additional information on the existing causal inference between the outcome and the treatment, which is rarely used. This fourth approach tackles a more general problem than estimating the counterfactual outcome and the treatment effect. A variant of the EM algorithm, introduced in~\cite{beji2020estimating}, used a parametric distribution to estimate the individual treatment effect. In the same way, the proposed CAE is a non-parametric method using a neural network architecture on auto-encoders to the causal hidden distribution.
Knowing the distribution of causal populations allows to classify individuals and predict the optimal treatments to assign. In addition to a simple prediction of counterfactual outcomes, this modelling provides a policy to support the treatments assignment for current individuals.
Thus, individuals for whom the treatment is beneficial, those for whom the treatment has no effect and those for whom it is noxious, can be targeted.

\section{Problem Setup}

Let $\mathcal{X} \subset \mathbb{R}^{d}$ be an $d$-dimensional continuous or discrete space and $\textbf{X}=\{\textbf{x}_i\}_{i=1}^n$ an i.i.d. sample, where each $\textbf{x}_i \in \mathcal{X}$ is the features. We note $y_{i,obs} \in \mathcal{Y}$ as the observed outcome and $t_i$ the treatment assignment of the individual $i \in \{1,..,n\}$. In this paper, we focus on a binary treatment $t_i \in \{0,1\}$. Individuals receiving treatment represent the \emph{test group} ($t_i=1$) and the others constitute the \emph{control group} ($t_i=0$). However, the model of causal population can be generalize to $K$-treatments. 
We indicate the outcome corresponding to the treatment value $t_k$ by $y_{t_k}$.
If a sample $\textbf{x}_i$ is treated ($t_i=1$), $y_{i,1}$ is observed and $y_{i,0}$ is missing. If not, $y_{i,0}$ and $y_{i,1}$ are respectively known and unknown. The potential outcome is defined as the concatenation of the observed outcome $y_{i,t_i}$ and the counterfactual outcomes $y_{i,(1-t_i)}$. Its estimation is necessary and sufficient to classify an individual in the four distinct causal populations: responders ($y_{i,0}=0$ and $y_{i,1}=1$), doomed ($y_{i,0}=0$ and $y_{i,1}=0$), survivors ($y_{i,0}=1$ and $y_{i,1}=1$) and anti-responders ($y_{i,0}=1$ and $y_{i,1}=0$).
We assume that the probability distribution of populations is generated by some random processes, involving a latent variable $z_i$. 

Two assumptions are required for the Rubin-Neyman causal model:
\begin{itemize}
    \item \emph{Overlap}. For all treatments $t_j \in \{0,1\}$ there is a non-zero probability of receiving the treatment, i.e
$0 < \mathbb{P}(t_j = 1|\textbf{x}) < 1$.
    \item \emph{Unconfoundedness}. Conditional on the features $\textbf{x}_i$, the potential outcome $y_i=\{y_{i,0},y_{i,1}\}$ is independent of the assigned treatment $t_i$.
\end{itemize} 
These two assumptions imply that observed outcomes and counterfactual outcomes can be treated in the same way and the distributions are exchanged.

Estimating the probability distribution of causal populations consists of two iterative steps: $z_i$ is generated from the posterior  distribution $q(z_i|\textbf{x}_i,y_{i,t_i})$ and $\textbf{x}_i$ is generated from the likelihood $p(\textbf{x}|z)$.

\section{Causal-Auto-Encoder (CAE)}

We propose a Causal-Auto-Encoder (CAE) that estimates the probability distribution of the causal populations, by reconstructing the covariates. In addition to the standard Auto-Encoder architecture, partial information, depending on the observed outcome and the assigned treatment, is applied on the latent space.

\subsection{Partial Information}

Causal constraints, introduced in~\cite{beji2020estimating}, based on the value of the treatment assignment and the observed outcome, introduces partial information on the latent distribution. They take the following form:
\begin{itemize}
    \item If $t_i=0$ and $y_{i,obs}=0$, then the individual $i$ is a responder or a doomed.
    \item If $t_i=0$ and $y_{i,obs}=1$, then the individual $i$ is a survivor or an anti-responder. 
    \item If $t_i=1$ and $y_{i,obs}=0$, then the individual $i$ is a doomed or an anti-responder.
    \item If $t_i=1$ and $y_{i,obs}=1$, then the individual $i$ is a responder or a survivor.
\end{itemize}
A prior information on the probabilities of the causal population can be introduced. For each individual $i \in \{1,..,n\}$, given $t_i$ and $y_{i,obs}$, two causal populations are by definition excluded. Therefore, their probability distribution can be considered as zero

\subsection{Global Architecture}

\input{figure_CAE}

The architecture is composed of three parts: an encoder, a mask built with the causal constraints and designed to set some units of the latent space to zero, and a decoder (see Figure~\ref{fig: Overview CAE} for an overview). 
The encoder takes as input the covariates $\textbf{x}_i$, the $i^{th}$ sample of the data for all $i \in \{1,..,n\}$. Each node of the initial layer represents a feature (a dimension of $\textbf{x}_i$). After compressing $\textbf{x}_i$ with a feedforward neural network, the latent variable $\mathbf{z}_i=\begin{pmatrix} z_{iR} & z_{iD} & z_{iS} & z_{iA} \end{pmatrix}$ is constrained by the mask $M(y_{i,obs},t_i)$. Each node $z_{ik}$ of the latent layer is assimilated to a causal population $k \in \{R,D,S,A\}$. $z_{ik}$ is activated or not (put to zero) according to the mask designed 
from the observed outcome $y_{i,obs}$ and the assigned treatment $t_i$. For example, if $t_i=0$ and $y_{i,obs}=0$ the individual is a survivor or an anti-responder. The designed mask is $M(y_{i,obs},t_i)^T= \begin{pmatrix} 0 & 0 & 1 & 1 \end{pmatrix}$ where the first coordinate (respectively the second, the third and the fourth) corresponds to the probabilities distribution of responders (respectively doomed, survivors and anti-responders).
The constrained latent variable $\tilde{\textbf{z}}_i$ is taken as input to the decoder. The decoder, built as a feedforward neural network, attempts to reconstruct the covariates. It outputs an estimated $\hat{\mathbf{x}}_i$ with the same size as the input covariates $\mathbf{x}_i$. The CAE is iteratively trained from a learning database $\{\textbf{x}_i\}_{i=1}^s$ (where $s$ is the number of sample used for learning), via back-propagation, minimizing the loss function $l_{CAE}(\mathbf{x}_i,\hat{\mathbf{x}}_i)$.

In the proposed CAE, the mask enforces some structured sparsity constraint on the hidden layer. An analogy can be seen with Sparse Auto-Encoders~\cite{ng2011sparse}, for which some units are deactivated , or a dropout structure~\cite{srivastava2014dropout}, which ignores (drop out) a certain set of neurons chosen at random during the training. These structure are used to improve the performance and reduce overfitting.
The significant difference remains in the structure of the sparsity or dropout constraint enforced in the CAE. In the proposed architecture, the structure is known and can be directly enforced without relying on a regularization scheme.

\subsection{Optimization}

The $m$-multiple layers encoder is modelled by a deterministic function:
\begin{equation}
    f_{\theta}(\textbf{x}) = s_1(W_1 s_2(W_2 \cdots s_m(W_m\textbf{x}+b_m)\cdots+b_2)+b_1)
\end{equation}
where in $\theta = \{W_1,W_2,W_3,..W_m,b_1,b_2,b_3,..,b_m\}$, $(W_1,W_2,W_3,..,W_m)$ are the weight parameters, $(b_1,b_2,b_3,..,b_m)$ the bias and are $(s_1,s_2,s_3,..,s_m)$ non-linear functions. 
To associate the last layer of the encoder to the probabilities distribution of the causal populations, we use the \emph{softmax(.)} activation function, defined as:
\begin{equation}
    u(z_{ik})=\frac{\exp(z_{ik})}{\sum_{j=1}^p \exp(z_{ij})}
\end{equation}{}
where $z_{ik}$ indicates a unit of the last latent layer of the encoder and $p$ the number of latent layer units.
The hidden representation is then mapped to construct an estimation of the input $\hat{\textbf{x}}$, using:
\begin{equation}
    \hat{\textbf{x}} = g_{\theta'}(\textbf{z}) =
    s'_1(W'_1 s'_2(W'_2  \cdots s'_m(W'_m\textbf{z}+b'_m)\cdots+b'_2)+b'_1)
\end{equation}
where $\theta' = \{W'_1,W'_2,W'_3,..W'_m,b'_1,b'_2,b'_3,..,b'_m\}$, are the parameters representing respectively the weights and the biases
and $(s'_1,s'_2,s'_3,..,s'_m)$ are non-linear functions. The parameters are optimized to minimize the loss function between the input $\textbf{x}$ and its reconstruction $\hat{\textbf{x}}$ over all training point. 

After computing the hidden representation and reconstructing the input, some hidden units are set to zero according to the prior information on the observed outcome and the assigned treatment. The mask $M(\textbf{y}_{obs},\textbf{t})$ is designed from all of the samples and inserted in the architecture as an auxiliary input. 
The hidden constrained representation is obtained by:
\begin{equation}
    \tilde{\textbf{z}}= M(y_{obs},t)^T \textbf{z}
\end{equation}
where each of the elements of the vector $M(y_{obs},t)$ correspond to a causal group. Assigning the first (also respectively the second, the third and the fourth) element to the group of responders (respectively doomed, survivors and anti-responders), the mask is obtained as:
\begin{equation}
    M(y_{i,obs},t_i)=
    \begin{pmatrix}
    \mathds{1}_{t_i=y_{i,obs}}\\
    1-y_{i,obs} \\
    y_{i,obs}\\
    \mathds{1}_{t_i \neq y_{i,obs}}
    \end{pmatrix}
\end{equation}

The set of parameters $(\theta,\theta')$ are optimized jointly via back-propagation. The reconstruction error estimator for this model is expressed as:
\begin{align}
\label{eq: loss_cae}
    l_{CAE}(\theta,\theta')
    & = l(\textbf{x},g_{\theta'}(M(y_{obs},t)^T f_{\theta}(\textbf{x})))
\end{align}
Using the standard mean square error, this optimisation problem boils down to minimize the following loss:
\begin{equation}
    l(\textbf{x},\hat{\textbf{x}})=||\hat{\textbf{x}}-\textbf{x}||^2_2 = ||g_{\theta'}(M(\textbf{y}_{obs},\textbf{t})^T f_{\theta}(\textbf{x}))-\textbf{x}||^2_2
    \label{eq: MSE_CAE}
\end{equation}

\subsection{Prediction}

Once the model is trained, the encoder block of the model is used to encode the probability distribution of causal populations $z_{ik}$.
Each individual is assigned to the causal population, for which the predicted probability is the highest. Moreover, the higher the probability of belonging to a causal population, the higher the confidence of the estimate.  
Knowing the causal population predicted for a given individual, its assigned treatment and its observed outcome, the counterfactual outcome can be directly estimated. 
As proven in ~\cite{beji2020estimating}, the Conditional Average Treatment Effect (CATE), defined as $\tau(\textbf{x}):=\mathbb{E}[Y_1-Y_0|X=\textbf{x}]$, can also be estimated from the probability distribution of the causal population $\tau(\textbf{x}_i) = (z_{iR} +z_{iS}) \mathbb{E}[\mathds{1}_{y_{i,1}=1}]
	  - (z_{iS}+ z_{iA})
	  \mathbb{E}[\mathds{1}_{y_{i,0}=1}]$.

\subsection{Latent Layer Architecture}

For the sake of simplicity, we have introduced this architecture with a single neuron to encode the probability of belonging to each causal group. 
This dimension reduction may be too drastic, as it compresses the information from the input space to a constrained four dimensional space.
To encode the four causal populations, we need to have at least four nodes, one for each population. We can also take more than one node per population. Intuitively, as the number of nodes increase, the more covariate information is preserved. The size of the mask depends on the number of nodes allocated to encode a population. It constraints therefore all the nodes associated to a causal population. The probability of belonging to a given population is calculated as the sum of the probabilities given by the nodes affected to this population.
In addition, we can add unconstrained nodes that only participate in encoding information from the covariates, without being disturbed by the causal priority. Their number can be chosen as large as required, depending on the number of features. However, the number of additional nodes must be lower than the size of the covariates, so that the information is distributed both on the additional nodes and on the distribution probability nodes. 

Note that no assumptions about the marginal or posterior distributions are requested. Conversely, we propose a method that works in the following two cases: (i) when the marginal likelihood $p(x)=\int q(z)p(\textbf{x}|z)dz$ is intractable and so the EM algorithm cannot be used; (ii) in large scale, when sampling methods as Monte Carlo EM are too expensive in term of batch optimization.

\section{Experiments}

In this section, in order to prove the efficiency of the proposed CAE architecture, we conduct an empirical study and compared the model with the state-of-the-art baselines.
Assessing the performance of a model on real-life data is challenging because the true counterfactual outcome is unknown and checking the accuracy of its estimate is unfeasible. 
Thus, we first conducted the experiments on semi-synthetic datasets. The data are constructed from real experimental or observational data, to which the counterfactual has been constructed synthetically from similar samples.
We then ran experiments on real-life datasets, for which we use a ranking metric that does not evaluate the estimation of the counterfactual outcome but the estimated CATE.
The architecture and the size of the latent space is discussed on all datasets.  

\subsection{Auto-Encoder Setting}

The encoder and the decoder are implemented as simple symmetric FeedForward networks. They are composed of four layers for which the dimension is halved or doubled at each layer. The size of the encoder input and the decoder output are given by the dimension of covariates. The dimension of the latent representation $\mathbf{z}$ depends on the architecture of the latent layer. In this experiments, four different architectures are chosen: $CAE_1$, $CAE_2$, $CAE_5$ and $CAE_{info5}$. The first (respectively the second and the third) used one (respectively two and five) nodes to encode the probability of each causal population. No additional unconstrained nodes is added in these models. The last architecture $CAE_{info5}$ is composed of one node to encode a causal population and five additional unconstrained nodes to capture the information of the features.

Each layer is followed by Batch Normalization, in order to stabilize the learning and to reduce the number of epochs required to train the model. By default, the activation function Rectified linear unit (ReLU), which returns the input value if it receives a positive input and zero otherwise $f(u) = u^{+}=max(0,u)$~\cite{glorot2011deep}, is used except for the last layer of the encoder. The softmax is chosen as activation function at the output of the encoder in order to obtain probabilities that sum up to one. Recall, that the mean squared error loss is used as loss function.
For all of the experiments, in order to update the parameters during the training phase, we use Adam optimizer, introduced in~\cite{kingma2014adam}, which is an adaption of the classical gradient descend.
It uses the average first and second order moments of the gradient and corrected the introduced bias with parameters that control the decay rates of these moving averages.

\subsection{Experimental Framework}

In this section, we compare the CAE with models using two classfiers (T-learner), one for the treatment and the other for the control group. To cover each of the categories, we build three baseline models using respectively logistic regressions (T-LR), regression trees (T-RF) and multi-layer perceptron classifiers (T-MLPC). 
These methods do not provide confidence intervals, but they show good performances in the literature~\cite{alaa2018limits}. 
On synthetic datasets, we also compared our model with the ECM algorithm~\cite{beji2020estimating} which has a strong prior on the distribution, since it fit an Gaussian mixture model on the causal populations.

We evaluate out-of-sample performance over $10$ trials and use a Wilcoxon signed-rank test with a level of $5\%$ to confirm the significance of the results.  Two metrics are considered: expected Precision in Estimation of Heterogeneous Effect (PEHE)~\cite{hill2011bayesian} and Area Under the Uplift Curve (AUUC)~\cite{diemert2018large}.
The expected PEHE is a metric for evaluating the performance of the counterfactual prediction. It is defined as the mean squared error between the true Individual Treatment Effect and its estimation:
\begin{equation}
    \epsilon_{PEHE} = \frac{1}{n} \sum_{i=1}^n (\mathbb{E}[y_{i,1}-y_{i,0} | \textbf{x}_i]-[\hat{y}_{i,1}-\hat{y}_{i,0}])^2.
\end{equation}
If the counterfactual outcome is unknown, as in a purely real dataset, this quantity cannot be calculated. The AUUC is used in \emph{Uplift} literature~\cite{betlei2018uplift,yamane2018uplift}. It is obtained by subtracting the respective Area Under Lift (AUL) curves on treatment and control groups. The lift curve represents the proportion of positive outcomes as a function of the percentage of individuals selected:
\begin{equation}
AUL_{\pi}(k)=\sum_{i=1}^k\sum_{d_i \in \pi(k)} \mathds{1}[y_i=1|t_i]-k\mathbb{E}[Y|T]
\end{equation} 
where $\pi$ is the ordering of the dataset satisfying $\hat{ITE}^{\pi}(x_i) \geq \hat{ITE}^{\pi}(x_j)$, $\forall i \leq j$, $\pi(k)$ the first $k$ samples sorted according to the descending ITE . 
The AUUC is obtained by cumulative summation:
\begin{multline}
AUUC 
\simeq \frac{1}{n} \sum_{k=1}^n AUUC_{\pi}(k)dk \\
= \frac{1}{n} \sum_{k=1}^n (AUL_{\pi}^{T=1}(k) - AUL_{\pi}^{T=0}(k))
\end{multline}
This metric provides a complementary evaluation because it assesses the estimation of the CATE of individuals w.r.t each other. It is a ranking evaluation measure which can be used even when the real counterfactual outcome is unavailable. 
It can be interpreted as the net gain in success rate provided that a given percentage of the population is treated according to the model.
This curve is remarkably similar to a ROC curve, and area under the uplift curve AUUC has comparable properties to the AUC. Note that the contrary to the ROC curve, the uplift curve is not always increasing. In the best model, the curve first increases (as the responder are treated) then remains constant (as the gain is null when doomed and survivors are treated) before it decreases with a negative gain as anti-responder are treated. The best performing method according to this criterion are the one with the largest AUUC.

\subsection{Results}

\subsubsection{Semi-Synthetic Datasets}

First, We conduce experiments on two semi-synthetic datasets: IHDP~\cite{hill2011bayesian} and Twins~\cite{louizos2017causal}.
The Infant Health and Development Program (IHDP) are randomized experiments that began in 1985 in the US designed to reduce the developmental and health problems of low birth weight premature infants. 
It contains $747$ instances with $25$ covariates ($6$ continuous, $19$ binary) measuring children characteristics like birth weight and head circumference and some features related to their mothers. The control and treatment groups are created according to the specialist visits. Children who received specialist visits are in the treatment group and the others in the control group.
The treatment and control groups are artificially imbalanced by removing a biased subset of the treated population ($608$ control, $139$ treated instances).
The dataset contains $49$ features ($21$ categorical, $28$ continuous) related to their mothers as if they smoked cigarettes, drank alcohol or took drugs. The sibling who have the bigger weight is considered as being treated. There are $32120$ instances with a balance between treated and untreated. The outcome variable corresponds to the mortality of each of the twins in their first year of life.
The real distribution of each causal population is unknown, but the potential outcomes is available.

\begin{table}[!t]
\begin{center}
\begin{tabular}{llcc}
\hline
&& $\epsilon_{PEHE}$ & AUUC \\
\hline  \vspace{0.1cm} \\
\multicolumn{2}{l}{IHDP} \vspace{0.2cm} \\

&$LR2$
& 0.63 +/- 0.06
& 2271 +/- 442
\\
&$RF2$
& 0.71 +/- 0.12 
& 2438 +/- 353
\\
&$MLPC2$
& 0.68+/- 0.04
& 2246 +/- 365
\\
\vspace{0.2cm}
&$CAE_1$
& 0.47 +/- 0.10
& \textbf{3535 +/- 481 ($\star$)}
\\
&$CAE_2$
& 0.69 +/- 0.24
& 2901 +/- 682
\\ 
&$CAE_5$
& 0.51 +/- 0.08
& 3178 +/- 359
\\ 
&$CAE_{info5}$
& \textbf{0.44 +/- 0.15 ($\star$)}
& 2989 +/- 860
\\  \vspace{0.1cm} \\
\multicolumn{2}{l}{Twins} \vspace{0.1cm} \\
&$LR2$
& 0.04 +/- 0.01
& 87 +/- 45 
\\
&$RF2$
& 0.04 +/- 0.01
& 86+/- 54
\\
&$MLPC2$
& \textbf{0.03 +/- 0.01}
& 97 +/- 60
\\
\vspace{0.2cm}
&$CAE_1$
& 0.35+/- 0.19
& \textbf{293 +/- 99 ($\star$)}
\\
&$CAE_2$
& 0.76+/- 0.20
& 185 +/- 159
\\ 
&$CAE_5$
& 0.51 +/- 0.25
& 119 +/- 83
\\ 
&$CAE_{info5}$
& 0.38 +/- 0.36
& 244 +/- 155 \vspace{0.1cm}
\\ \hline 
\end{tabular}
\end{center}
\caption{Numerical results on CAE with different structures of latent variables compared to baselines on the semi-synthetic datasets. A ($\star$) indicates a significant result using a Wilcoxon signed-rank test at level of $5\%$ compared to second best baseline. The best performing model is the one with the lowest $\epsilon_{pehe}$ and the highest $AUUC$.}
\label{tab:results_semi_synth}
\end{table}

Results in Table~\ref{tab:results_semi_synth} show the high performance of the CAE compared to baselines. Because of the complexity of distributions, $\epsilon_{pehe}$ values on IHDP dataset are high for all models compared to those on Twins, but we see that $CAE_{info5}$ and $CAE_1$ are the most efficient.
On twins dataset, the present methods allow a more accurate prediction of the counterfactual, however the ranking of individuals in relation to each other is better with $CAE$. Although the $CAE_1$ is the best model for ranking individuals according to their ITE, the $CAE_5$ is one of CAE's weakest $\epsilon_{pehe}$ models because this model captures more information on the initial structure of the data, without any perturbation from causality constraints.

\subsubsection{Real Life Datasets}

\begin{figure*}[!ht]

    \centering
    \begin{subfigure}[t]{0.45\textwidth}
        \centering
        \includegraphics[width=\linewidth]{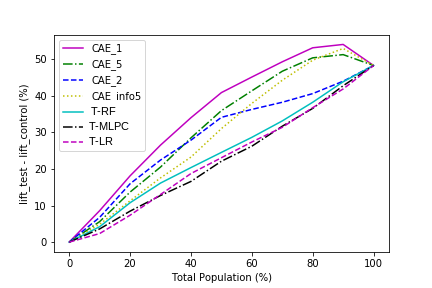} 
        \caption{IHDP} \label{figure:UC_CAE_IHDP}
    \end{subfigure}
    \hfill
    \begin{subfigure}[t]{0.45\textwidth}
        \centering
        \includegraphics[width=\linewidth]{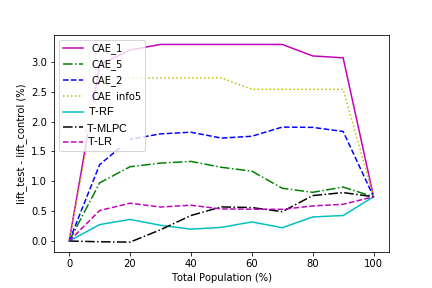} 
        \caption{Twins} \label{figure:UC_CAE_Twins}
    \end{subfigure}
    \begin{subfigure}[t]{0.45\textwidth}
        \centering
        \includegraphics[width=\linewidth]{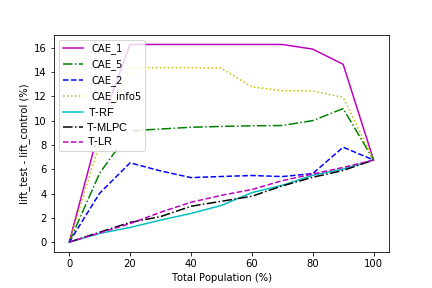} 
        \caption{Email} 
        \label{figure:UC_CAE_Email}
    \end{subfigure} 
    \hfill 
    \begin{subfigure}[t]{0.45\textwidth}
        \centering
        \includegraphics[width=\linewidth]{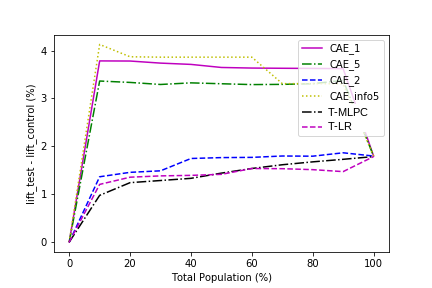} 
       \caption{Criteo} 
       \label{figure:UC_CAE_Criteo}
    \end{subfigure}
    \caption{Uplift curves for real (purely real and semi-synthetic) datasets, built as the difference of the lift curves on the treatment and control groups.}
    \label{figure:real datasets Uplift curve}
    \end{figure*}

In absence of groundtruth for the ITE on real data, we observe the behavior of the uplift curve, on two datasets: Email~\cite{hillstrom2008minethatdata} and Criteo~\cite{diemert2018large} (see Figure~\ref{figure:real datasets Uplift curve}). 
Email dataset contains $64 000$ customers divided in two groups: one receives an email from a marketing campaign ($67\%$), the treatment group, and the other does not ($34\%$), the control group. After two weeks, customers behaviors i.e. visiting the website or not, used as outcome, were tracked according to the customers features ($6$ categorical, $1$ continuous).
Criteo is a dataset that collects information on consumer behaviour after being exposed or not to an advertisement.
It contains $25$M units with $12$ continuous anonymized features, a treatment indicator and $2$ binary outcomes (visits and conversions). The treatment group is composed of $84.6\%$ of individuals and control group of $15,4\%$. 
In this two dataset, the counterfactual outcome is not available.

Compared to the baselines, our algorithms perform better on both Email and Criteo datasets. Among $CAE$ models, the $CAE_1$ uplift curve is the closest to the optimal model, followed by the $CAE_{info5}$. 
Note that the larger the dataset, the closer the uplift curve is to the perfect form.
However, the $RF2$ model has a training time that disqualifies it from the other models, in large scale setups.

\section{Conclusion}

In this paper, we propose an auto-encoder architecture which by means of partial information on latent variables, estimates the probability distribution of causal populations. This method provides a prediction of counterfactuals outcomes and identifies the combination of treatments that maximizes the global effect. The proposed model is efficient on real data and has the significant advantage to be non-parametric, interpretable (as it identifies the causal populations) and applicable on a large scale datasets. Although the synthetic data highlighted the limits of our model in low complexity, the structure can be easily adapted to a particular situation. Also, this method is versatile enough to work under a multi-treatment setup, poorly covered in the literature. However, in that case, the question of the correlation between treatments remains unresolved.

\bibliographystyle{apalike}
\bibliography{biblio}

\end{document}

%% file: figure_CAE.tex
\begin{figure*}[!ht]
\begin{center}
\begin{tikzpicture}[thick,scale=1]
\tikzstyle{node2}=[circle, draw, inner sep=2pt, minimum width=3pt, minimum size=0.2cm, fill=gray]
\tikzstyle{node}=[circle, fill=white, draw, inner sep=2pt, minimum width=3pt, minimum size=0.2cm]

\node (X) at (0,4) {$\mathbf{x}_i$};

\foreach \i in {1,...,7}{
        \node[node] (a\i) at (0, 0.5*\i) {};
}
\draw [<->] (-0.2,0.3) --(-0.2,3.7) node[above,midway, rotate=90]{$d=dim(\mathbf{x}_i)$};

\node (M) at (0,0) { $M(y_{i,obs},t_i)$};
\node[node2] (m1) at (0, -0.5*1) {};
\node[node] (m2) at (0, -0.5*2) {};
\node[node2] (m3) at (0, -0.5*3) {};
\node[node] (m4) at (0, -0.5*4) {};



\node (Z) at (4,3.5) {$\mathbf{z}_i$};

\node[node] (c1) at (4, 1.2) {};
\node[node] (c2) at (4, 1.7) {};
\node[node] (c3) at (4, 2.2) {};
\node[node] (c4) at (4, 2.7) {};

\foreach \i in {1,...,7}{
    \foreach \j in {1,...,4}{
        \draw (a\i) edge[gray!50] (c\j);
    }
}


\node[draw, minimum width=3cm, minimum height=1.5cm, fill=white, rounded corners=3pt, rotate=90] at (2,2) {\Large Encoder};

\node (ZM) at (7,3.5) {$\tilde{\mathbf{z}}_i$};

\node[node] (d1) at (7,1.2) {};
\node[node2] (d2) at (7,1.7) {};
\node[node] (d3) at (7,2.2) {};
\node[node2] (d4) at (7,2.7) {};

\draw (c1) edge[gray!50] (d1);
\draw (c2) edge[gray!50] (d2);
\draw (c3) edge[gray!50] (d3);
\draw (c4) edge[gray!50] (d4);

\draw[draw=gray!50] (m1) -- (4, -0.5*1) node[midway]{\textit{Responder}} -- (d4);
\draw[draw=gray!50] (m2) -- (4, -0.5*2) node[midway]{\textit{Doomed}} -- (d3);
\draw[draw=gray!50] (m3) -- (4, -0.5*3) node[midway]{\textit{Survivor}} -- (d2);
\draw[draw=gray!50] (m4) -- (4, -0.5*4) node[midway]{\textit{Anti-responder}} -- (d1);


\node (Xr) at (11,4) {$\hat{\mathbf{x}}_i$};

\foreach \i in {1,...,7}{
        \node[node] (f\i) at (11, 0.5*\i) {};
}

\foreach \i in {1,...,4}{
    \foreach \j in {1,...,7}{
        \draw (d\i) edge[gray!50] (f\j);
    }
}

\draw [<->] (11.2,0.3) --(11.2,3.7) node[below,midway, rotate=90]{$d$};

\node[draw, minimum width=3cm, minimum height=1.5cm, fill=white, rounded corners=3pt, rotate=90] at (9,2) {\Large Decoder};

\draw[<->,thick] (X) -- (0,4.5) -- (11,4.5) node[above,midway]{$l_{CAE}(\mathbf{x}_i,\hat{\mathbf{x}}_i)$} --(Xr);

\node[node, label=right:Active nodes] at (8,-1) {};
\node[node2, label=right:Inactive nodes] at (8,-1.5) {};

\end{tikzpicture}
\end{center}

\caption{Overview of the Causal-Auto-Encoder (CAE) architecture. White nodes (respectively gray nodes) correspond to active neurons (respectively inactive neurons). $\mathbf{x}_i$ are the covariates and $\hat{\mathbf{x}}_i$ is its reconstruction by the Auto-Encoder. $\mathbf{z}_i$ is the latent variable constructed by the encoder. It is then constrained by the mask $M(y_{i,obs},t_i)$ (some of this neurons are deactivate), from $y_{i,obs}$ the observed outcome and $t_i$ the treatment assignment.} 
\label{fig: Overview CAE}
\end{figure*}

%% file: main.bbl
\begin{thebibliography}{}

\bibitem[Alaa and Schaar, 2018]{alaa2018limits}
Alaa, A. and Schaar, M. (2018).
\newblock Limits of estimating heterogeneous treatment effects: Guidelines for
  practical algorithm design.
\newblock In {\em ICML}.

\bibitem[Alaa and van~der Schaar, 2017]{alaa2017bayesian}
Alaa, A.~M. and van~der Schaar, M. (2017).
\newblock Bayesian inference of individualized treatment effects using
  multi-task gaussian processes.
\newblock In {\em Advances in Neural Information Processing Systems}, pages
  3424--3432.

\bibitem[Alaa et~al., 2017]{alaa2017deep}
Alaa, A.~M., Weisz, M., and Van Der~Schaar, M. (2017).
\newblock Deep counterfactual networks with propensity-dropout.
\newblock {\em arXiv preprint arXiv:1706.05966}.

\bibitem[Beji et~al., 2020]{beji2020estimating}
Beji, C., Bon, M., Yger, F., and Atif, J. (2020).
\newblock Estimating individual treatment effects through causal populations
  identification.
\newblock {\em arXiv preprint arXiv:2004.05013}.

\bibitem[Betlei et~al., 2018]{betlei2018uplift}
Betlei, A., Diemert, E., and Amini, M.-R. (2018).
\newblock Uplift prediction with dependent feature representation in imbalanced
  treatment and control conditions.
\newblock In {\em International Conference on Neural Information Processing},
  pages 47--57. Springer.

\bibitem[Diemert et~al., 2018]{diemert2018large}
Diemert, E., Betlei, A., Renaudin, C., and Amini, M.-R. (2018).
\newblock A large scale benchmark for uplift modeling.
\newblock In {\em Proceedings of the AdKDD and TargetAd Workshop, KDD}. ACM.

\bibitem[Foster et~al., 2011]{foster2011subgroup}
Foster, J.~C., Taylor, J.~M., and Ruberg, S.~J. (2011).
\newblock Subgroup identification from randomized clinical trial data.
\newblock {\em Statistics in medicine}, 30(24):2867--2880.

\bibitem[Glorot et~al., 2011]{glorot2011deep}
Glorot, X., Bordes, A., and Bengio, Y. (2011).
\newblock Deep sparse rectifier neural networks.
\newblock In {\em Proceedings of the fourteenth international conference on
  artificial intelligence and statistics}, pages 315--323. JMLR Workshop and
  Conference Proceedings.

\bibitem[Hill, 2011]{hill2011bayesian}
Hill, J.~L. (2011).
\newblock Bayesian nonparametric modeling for causal inference.
\newblock {\em Journal of Computational and Graphical Statistics},
  20(1):217--240.

\bibitem[Hillstrom, 2008]{hillstrom2008minethatdata}
Hillstrom, K. (2008).
\newblock The minethatdata e-mail analytics and data mining challenge.
\newblock {\em MineThatData blog}.

\bibitem[Hitsch and Misra, 2018]{hitsch2018heterogeneous}
Hitsch, G.~J. and Misra, S. (2018).
\newblock Heterogeneous treatment effects and optimal targeting policy
  evaluation.
\newblock {\em Available at SSRN 3111957}.

\bibitem[Johansson et~al., 2016]{johansson2016learning}
Johansson, F., Shalit, U., and Sontag, D. (2016).
\newblock Learning representations for counterfactual inference.
\newblock In {\em International conference on machine learning}, pages
  3020--3029.

\bibitem[Kang et~al., 2007]{kang2007demystifying}
Kang, J.~D., Schafer, J.~L., et~al. (2007).
\newblock Demystifying double robustness: A comparison of alternative
  strategies for estimating a population mean from incomplete data.
\newblock {\em Statistical science}, 22(4):523--539.

\bibitem[Kingma and Ba, 2014]{kingma2014adam}
Kingma, D.~P. and Ba, J. (2014).
\newblock Adam: A method for stochastic optimization.
\newblock {\em arXiv preprint arXiv:1412.6980}.

\bibitem[Knaus et~al., 2021]{knaus2021machine}
Knaus, M.~C., Lechner, M., and Strittmatter, A. (2021).
\newblock Machine learning estimation of heterogeneous causal effects:
  Empirical monte carlo evidence.
\newblock {\em The Econometrics Journal}, 24(1):134--161.

\bibitem[K{\"u}nzel et~al., 2019]{kunzel2019metalearners}
K{\"u}nzel, S.~R., Sekhon, J.~S., Bickel, P.~J., and Yu, B. (2019).
\newblock Metalearners for estimating heterogeneous treatment effects using
  machine learning.
\newblock {\em Proceedings of the national academy of sciences},
  116(10):4156--4165.

\bibitem[Louizos et~al., 2017]{louizos2017causal}
Louizos, C., Shalit, U., Mooij, J.~M., Sontag, D., Zemel, R., and Welling, M.
  (2017).
\newblock Causal effect inference with deep latent-variable models.
\newblock In {\em NIPS}.

\bibitem[Lu et~al., 2018]{lu2018estimating}
Lu, M., Sadiq, S., Feaster, D.~J., and Ishwaran, H. (2018).
\newblock Estimating individual treatment effect in observational data using
  random forest methods.
\newblock {\em Journal of Computational and Graphical Statistics},
  27(1):209--219.

\bibitem[Ng et~al., 2011]{ng2011sparse}
Ng, A. et~al. (2011).
\newblock Sparse autoencoder.
\newblock {\em CS294A Lecture notes}, 72(2011):1--19.

\bibitem[Nie and Wager, 2020]{nie2020quasi}
Nie, X. and Wager, S. (2020).
\newblock Quasi-oracle estimation of heterogeneous treatment effects.
\newblock {\em Biometrika, forthcoming}.

\bibitem[Rubin, 1974]{Rubin_1974}
Rubin, D. (1974).
\newblock Estimating causal effects of treatments in randomized and
  nonrandomized studies.
\newblock {\em Journal of Educational Psychology}, 66(5):688--701.

\bibitem[Rubin, 1975]{rubin1975bayesian}
Rubin, D.~B. (1975).
\newblock Bayesian inference for causality: The importance of randomization.
\newblock In {\em The Proceedings of the social statistics section of the
  American Statistical Association}, volume 233, page 239. American Statistical
  Association Alexandria, VA.

\bibitem[Shalit et~al., 2017]{shalit2017estimating}
Shalit, U., Johansson, F.~D., and Sontag, D. (2017).
\newblock Estimating individual treatment effect: generalization bounds and
  algorithms.
\newblock In {\em Proceedings of the 34th International Conference on Machine
  Learning-Volume 70}, pages 3076--3085. JMLR. org.

\bibitem[Srivastava et~al., 2014]{srivastava2014dropout}
Srivastava, N., Hinton, G., Krizhevsky, A., Sutskever, I., and Salakhutdinov,
  R. (2014).
\newblock Dropout: a simple way to prevent neural networks from overfitting.
\newblock {\em The journal of machine learning research}, 15(1):1929--1958.

\bibitem[Yamane et~al., 2018]{yamane2018uplift}
Yamane, I., Yger, F., Atif, J., and Sugiyama, M. (2018).
\newblock Uplift modeling from separate labels.
\newblock In {\em Advances in Neural Information Processing Systems}.

\bibitem[Yoon et~al., 2018]{yoon2018ganite}
Yoon, J., Jordon, J., and van~der Schaar, M. (2018).
\newblock Ganite: Estimation of individualized treatment effects using
  generative adversarial nets.
\newblock In {\em ICLR}.

\end{thebibliography}
